\documentclass[runningheads]{llncs}
\usepackage{graphicx}
\usepackage{tikz}
\usepackage{comment}
\usepackage{amsmath,amssymb} % define this before the line numbering.
\usepackage{color}
\usepackage{multirow}
\usepackage[accsupp]{axessibility}  % Improves PDF readability for those with disabilities.
\usepackage{subcaption}
\usepackage{wrapfig}
\usepackage{xspace}
\makeatletter
\DeclareRobustCommand\onedot{\futurelet\@let@token\@onedot}
\def\@onedot{\ifx\@let@token.\else.\null\fi\xspace}
\def\eg{\emph{e.g}\onedot} 
\def\ie{\emph{i.e}\onedot}

\begin{document}

\pagestyle{headings}
\mainmatter

\title{Decoupled Adversarial Contrastive Learning \\ for Self-supervised Adversarial Robustness}

% INITIAL SUBMISSION 
% \begin{comment}
% \titlerunning{ECCV-22 submission ID \ECCVSubNumber} 
% \authorrunning{ECCV-22 submission ID \ECCVSubNumber} 
% \author{Anonymous ECCV submission}
% \institute{Paper ID \ECCVSubNumber}
% \end{comment}
%******************

% CAMERA READY SUBMISSION
%\begin{comment}
\titlerunning{DeACL: Decoupled Adversarial Contrastive Learning}
% If the paper title is too long for the running head, you can set
% an abbreviated paper title here
%
% \author{Chaoning Zhang\inst{1}\orcidID{0000-1111-2222-3333}
\author{
Chaoning Zhang$^{*}$\inst{1} \and
Kang Zhang$^{*}$\inst{1} \and
Chenshuang Zhang\inst{1} \and
Axi Niu\inst{2} \and
Jiu Feng\inst{3} \and
Chang D. Yoo\inst{1} \and
In So Kweon\inst{1} }
\authorrunning{C. Zhang et al.}
% First names are abbreviated in the running head.
% If there are more than two authors, 'et al.' is used.
%
\institute{
Korea Advanced Institute of Science and Technology (KAIST), Daejeon, Korea 
\email{chaoningzhang1990@gmail.com, zhangkang@kaist.ac.kr} \and
Northwestern Polytechnical University, Xi'an, China  \and 
Sichuan University, Chengdu, China
}
\def\thefootnote{*}\footnotetext{Equal Contribution.}\def\thefootnote{\arabic{footnote}}

%******************
\maketitle

\begin{abstract}

\textit{Adversarial training} (AT) for robust representation learning and \textit{self-supervised learning} (SSL) for unsupervised representation learning are two active research fields. Integrating AT into SSL, multiple prior works have accomplished a highly significant yet challenging task: learning robust representation without labels. A widely used framework is adversarial contrastive learning which couples AT and SSL, and thus constitutes a very complex optimization problem. Inspired by the divide-and-conquer philosophy, we conjecture that it might be simplified as well as improved by solving two sub-problems: non-robust SSL and pseudo-supervised AT. This motivation shifts the focus of the task from seeking an optimal integrating strategy for a coupled problem to finding sub-solutions for sub-problems. With this said, this work discards prior practices of directly introducing AT to SSL frameworks and proposed a two-stage framework termed \underline{De}coupled \underline{A}dversarial \underline{C}ontrastive \underline{L}earning (DeACL). Extensive experimental results demonstrate that our DeACL achieves SOTA self-supervised adversarial robustness while significantly reducing the training time, which validates its effectiveness and efficiency. Moreover, our DeACL constitutes a more explainable solution, and its success also bridges the gap with semi-supervised AT for exploiting unlabeled samples for robust representation learning. The code is publicly accessible at \url{https://github.com/pantheon5100/DeACL}. 

\end{abstract}

\setlength{\intextsep}{0pt}
\section{Introduction}

Despite the phenomenal success in a wide range of applications~\cite{he2016deep,huang2017densely,zhang2019revisiting,zhang2020resnet}, deep neural networks (DNNs) are widely recognized to be vulnerable to adversarial examples~\cite{szegedy2013intriguing,goodfellow2014explaining}. Adversarial training (AT) and its variants have become the de facto standard approach for learning an adversarially robust model~\cite{madry2017towards,zhang2019theoretically}. AT targets robust generalization~\cite{schmidt2018adversarially} which requires more data than standard training. In practice, however, samples with ground-truth (GT) labels are much more difficult to obtain than their unlabeled counterparts. To partly or fully remove the dependence on human annotation, unlabeled samples can be exploited for learning robust representation. 

Multiple works~\cite{uesato2019labels,carmon2019unlabeled,zhai2019adversarially,najafi2019robustness} have independently shown that unlabeled samples improve adversarial robustness in the semi-supervised setting. The performance of such semi-supervised AT, however, is often reported to be poor when only a small amount of labelled samples are available. Therefore, an interesting question is whether reasonable robustness can be achieved with \textit{only} unlabeled samples. The past few years have witnessed substantial progress in the field of self-supervised learning (SSL)~\cite{chen2020simple,he2019moco,chen2021exploring} for representation learning without GT labels. Inspired by such progress, multiple works~\cite{xu2020adversarial,jiang2020robust,fan2021does,kim2020adversarial} have shown the success of adversarial contrastive learning (CL) for achieving robustness without labels, which constitutes a positive answer to the above question.

Nonetheless, robust SSL has been often recognized as a challenging problem due to its two mixed challenging goals: (a) \textit{unsupervised} representation learning; (b) \textit{robust} representation learning. The first goal can be readily realized by SOTA SSL frameworks, such as contrastive learning (CL)-based SimCLR~\cite{chen2020simple}, MoCo~\cite{he2020momentum}, while AT constitutes a go-to solution for the second goal. Thus, a line of works~\cite{xu2020adversarial,kim2020adversarial,jiang2020robust,fan2021does} choose a natural strategy by introducing AT into SimCLR or MoCo to perform adversarial CL. Despite having such off-the-shelf solutions for both SSL and AT, how to effectively integrate the two techniques as an optimal solution remains not fully clear. Searching for such an optimal combining strategy is non-trivial because the two goals are entangled in the optimization. Moreover, SSL and AT often require different configuration choices for their respective goals, and combining them inevitably involves a trade-off between them. Inspired by the design philosophy of the divide-and-conquer algorithm, we conjecture that the task might be simplified by solving two sub-problems in a decoupled manner. This frustratingly simple motivation brings a fundamental shift for the focus of robust SSL: from \textit{seeking an optimal combining strategy for a coupled problem} to \textit{finding sub-solutions for sub-problems}.

\begin{wrapfigure}[14]{r}{0.45\textwidth}
  \begin{center}
    \includegraphics[width=0.48\textwidth]{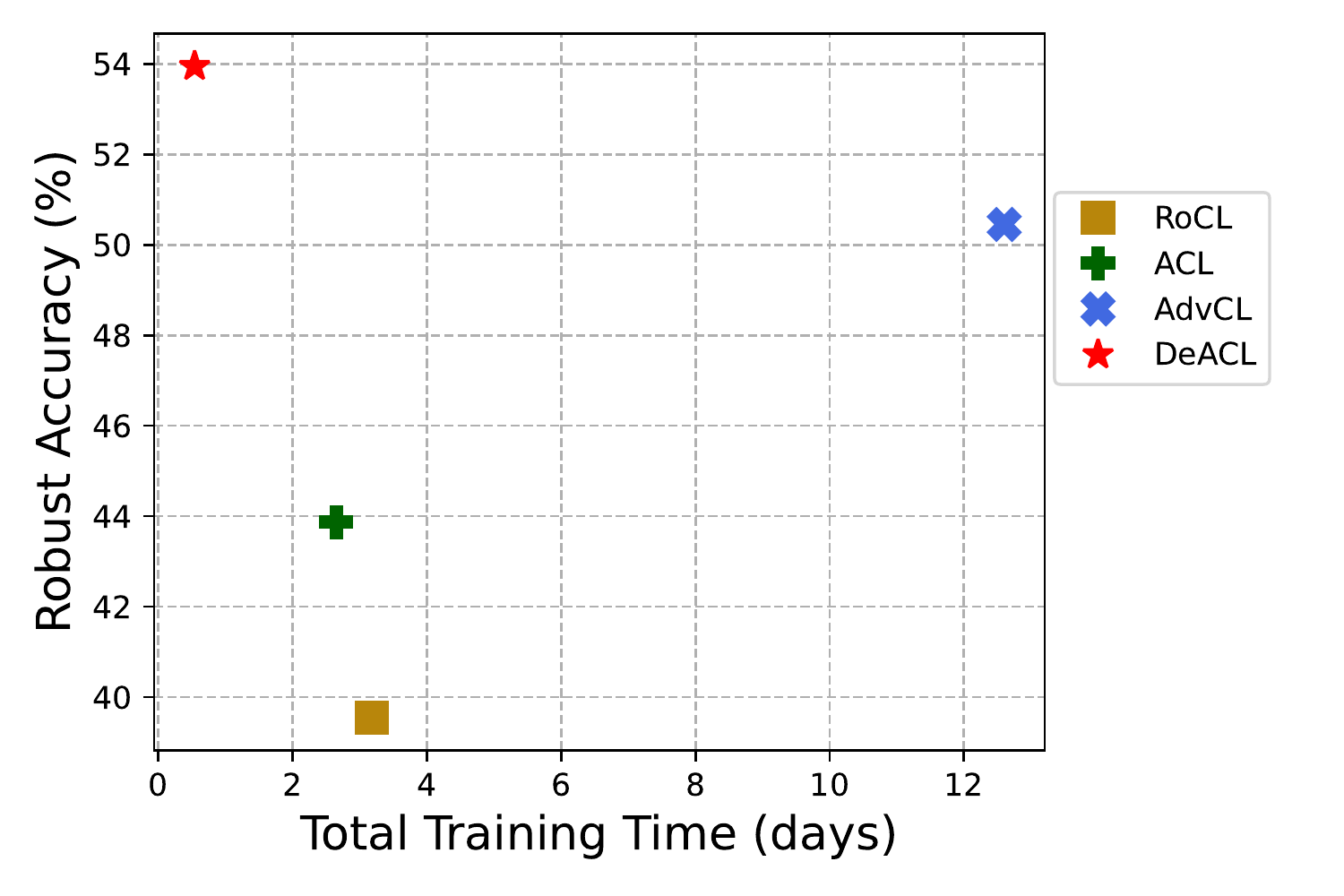}
  \end{center}
  \caption{Comparison of different methods on robust accuracy and total training time.}
  \label{fig:teaser}
\end{wrapfigure}
To this end, this work discards the prior practice~\cite{xu2020adversarial,kim2020adversarial,jiang2020robust,fan2021does} of introducing AT to SSL frameworks and proposes a new two-stage framework termed \underline{De}coupled \underline{A}dversarial \underline{C}ontrastive \underline{L}earning (DeACL). At stage 1, we perform standard (\ie\ non-robust) SSL to learn instance-wise representation as a target vector. At stage 2, the obtained target vectors can be used for facilitating AT in a \textbf{pseudo-supervised} manner for learning robust representation. We find that DeACL significantly benefits from the configuration for second-stage AT being set differently from that of first-stage SSL.

Except for enabling flexible yet simple configuration choices, another important side benefit of our DeACL is to require much fewer computation resources. At first sight, it might be counter-intuitive that two-stage approaches can be much faster than single-stage counterparts. Ignoring specific implementation details, the basic rationale is as follows. SSL typically requires $M$ times more iterations than their supervised counterpart, and AT is often $N$ times slower than standard training. Compared with supervised standard training, combining them into a single-stage makes it $M\times N$ times slower, while our DeACL makes it only $M+N$ times slower due to the disentangling effect. As shown in Figure~\ref{fig:teaser}, with SimCLR as the baseline SSL, our DeACL achieves state-of-the-art robustness while significantly reducing the required training time. The superior performance of our approach is also confirmed under adversarial full fine-tuning.

Overall, this work studies self-supervised robust representation learning. We summarize the contributions as follows:

\begin{itemize}
    \item In contrast to existing works seeking an optimal strategy for combing SSL and AT to achieve robust SSL, our work investigates a different approach by solving two sub-problems in a divide-and-conquer manner, which yields a novel two-stage DeACL framework for robust SSL.
    \item The proposed DeACL has two advantages: (a) enabling flexible configuration for the two sub-problems; (b) requiring much fewer computation resources. Extensive experiments demonstrate that DeACL achieves SOTA robustness while significantly reducing the training time. 
    \item Our DeACL also constitutes a more explainable solution for robust SSL and its success also bridges the gap with semi-supervised AT for exploiting unlabeled samples for robust representation learning. 
\end{itemize}

\section{Related works}
The task of robust SSL lies in the intersection between SSL and AT to learn robust feature representation without GT labels. SSL and AT are two active research fields, for which we summarize their recent progress.

\textbf{Development in SSL.} The success of SSL has been demonstrated in a wide range of applications, ranging from natural language processing \cite{Lan2020ALBERT,radford2019language,devlinetal2019bert,su2020vlbert,nie2020dc} to more recent vision tasks \cite{li2021esvit,chen2021mocov3,el2021xcit}. Without the need for GT labels annotated by the human, early SSL approaches leverage handcrafted ``pretext" tasks, like solving Jigsaw puzzle~\cite{gidaris2018unsupervised} or predicting image rotation~\cite{noroozi2016unsupervised}, while recent methods seek to learn augmentation-invariant representation~\cite{bachman2019learning,he2020momentum,chen2020simple,caron2020unsupervised,grill2020bootstrap}. To make the encoder augmentation-invariant, a commonly adopted practice is to minimize the distance between a positive pair, \ie\ two views augmented from the same image based on a Siamese network architecture. A widely known issue in SSL is that the network might output an undesired constant, for which contrastive learning (CL) provides a satisfactory solution by maximizing the distance between negative samples, \ie\ views of different images. CL has been widely investigated in ~\cite{oord2018representation,hjelm2018learning,wu2018unsupervised,zhuang2019local,bachman2019learning,henaff2020data,tian2020contrastive,chen2020simple,he2020momentum,wang2020understanding,wang2020DenseCL,yeh2021decoupled}, contributing to the progress of SSL. Recently, multiple works~\cite{chen2021exploring,grill2020bootstrap,ermolov2021whitening,zbontar2021barlow,bardes2021vicreg} have also explored non-contrastive SSL. A unified perspective on contrasitve and non-contrastive SSL is provided in ~\cite{zhang2022how,zhang2022dual}.

\textbf{Development in AT.} To improve adversarial robustness, early works have attempted with various image processing or detection techniques, most of which, however, are found to give a false sense of robustness~\cite{carlini2017adversarial,athalye2018obfuscated,croce2020reliable}. Currently, AT and its variants are widely recognized as powerful solutions to improve model robustness, among which Mardy-AT~\cite{madry2017towards} and Trades-AT~\cite{zhang2019theoretically} are two widely used baselines. From the perspective of model architecture, AT often requires a larger model capacity~\cite{uesato2019labels,xie2019intriguing}. Moreover,~\cite{xie2020smooth,pang2020bag} have found that a smooth activation function, like parametric softplus, is often but not always~\cite{gowal2020uncovering} helpful for AT. From the perspective of tricks,~\cite{pang2020bag} has performed a comprehensive evaluation for bags of tricks in AT and found that most of them provide no or trivial performance boost over Mardy-AT and Trades-AT if basic hyperparameters, such as weight decay, are set to proper values. From the perspective of data, ~\cite{uesato2019labels,carmon2019unlabeled,zhang2019bottleneck} have shown that unlabeled data can be helpful for robustness improvement over a basic supervised baseline. However, those approaches still depend on a large amount of labeled samples. For example,~\cite{uesato2019labels,carmon2019unlabeled,zhai2019adversarially} have shown that robust accuracy drops significantly when only 10\% of the CIFAR10 labels are available. Universal AT~\cite{benz2021universal} has also been investigated for defending against universal adversarial perturbations~\cite{moosavi2017universal,zhang2020understanding,benz2020double,zhang2021universal,zhang2021data}.

\textbf{Self-supervised adversarial robustness.} Clearly, self-supervised AT, \ie\ achieving robustness with only unlabeled samples, can be even more challenging than the semi-supervised AT setting. Nonetheless, multiple recent works~\cite{chen2020adversarial,jiang2020robust,kim2020adversarial,gowal2020self,fan2021does} have demonstrated encouraging success in this challenging yet highly significant direction. Prior attempts mainly focused on finding effective techniques to combine SSL and AT. 
What differentiates our approach from prior attempts~\cite{xu2020adversarial,jiang2020robust,kim2020adversarial,gowal2020self,fan2021does} lies in disentangling robust SSL into two decoupled sub-problems (SSL and AT) which can be solved in two stages. In the following section, we will detail existing single-stage frameworks as well as the motivation behind our two-stage framework. 

\section{Proposed method}
To avoid ambiguity, we start by presenting the problem of our interest, \ie\ robust SSL, and common fine-tuning methods for evaluating the learned robust representation. Then, we briefly summarize how prior attempts~\cite{chen2020adversarial,jiang2020robust,kim2020adversarial,gowal2020self,fan2021does} solve this problem in a single-stage framework. Compared with standard supervised training, either SSL or AT makes the optimization more complex, while simultaneously realizing SSL and AT clearly makes the problem complexity to an even higher level thus is difficult to solve. Inspired by the philosophy of the divide-and-conquer algorithm, we divide the complex robust SSL problem into two sub-problems: non-robust SSL and pseudo-supervised AT, and sequentially conquer them. We identify multiple important details that need to be configured differently for AT at stage 2 from standard SSL at stage 1.

\subsection{Problem statement}

\textbf{Robust SSL.} The goal of robust SSL is to learn robust feature representation with only unlabeled samples so that the model can be trained by a self-supervision loss, such as InfoNCE in CL-based SSL frameworks~\cite{he2020momentum,chen2020simple}. Note that this is different from a semi-supervised setting, where labeled samples are used together with unlabeled dataset. By contrast, robust SSL \textit{exclusively only} utilizes unlabeled dataset. 

\textbf{Standard linear finetuning.} For quantitatively evaluating the learned representation, a common practice is to train a linear classifier (denoted as $\phi_{\theta_{c}}$) on top of the pretrained encoder (denoted as $f_{\theta_e}$) as:

\begin{equation}
    \text{SLF:}~\underset{\theta_{c} }{min} \mathbb{E}_{(x,y)\in \mathcal{D}} \ell _{CE}( \phi _{\theta _{c} } \circ f_{\theta_{e} }(x),y ) ,
    \label{eq:slf}
\end{equation}
where $\ell _{CE}$ represents the supervised CE loss with GT-labels $y$ over a certain dataset $\mathcal{D}$. This is often termed \textit{standard linear finetuning} (SLF)~\cite{jiang2020robust,fan2021does} since only a linear classifier is updated on Clean Examples (CEs). Training such a linear classifier allows access to the ground-truth labels; otherwise, the learned representation in the encoder cannot be evaluated. To not break the rule of the SSL task, the backward gradient can only be propagated to the linear classifier so that the pretrained encoder is fixed during the evaluation. The quality of learned robust representation is finally evaluated on the full model $\phi _{\theta _{c}} \circ f_{\theta_e}$ by measuring its robust accuracy under PGD attack~\cite{madry2017towards} or autoattack~\cite{croce2020reliable}. 
\textbf{Adversarial full finetuning.} Except for the above linear finetuning as the primary evaluation metric, one can also make the constraint less strict in the finetuning stage to perform adversarial fullfinetuning~\cite{kim2020adversarial} (AFF). AFF allows the encoder to be updated during the finetunning as: 

\begin{equation}
    \text{AFF:}~\underset{\theta_{c}, \theta_{e}}{min} \mathbb{E}_{(x,y)\in \mathcal{D}} \ell _{CE}( \phi _{\theta _{c} } \circ f_{\theta_{e} }(x + \delta),y ),
    \label{eq:aff}
\end{equation}
where $\sigma$ is adversarial perturbation. It is worth highlighting that the weight initialization from robust SSL significantly improves the convergence speed of supervised AT together with a non-trivial performance boost. Note that we do not consider standard full finetuning because it cannot generate a robust model.

\textbf{Basic setup.} Following~\cite{kim2020adversarial,jiang2020robust,fan2021does}, we adopt ResNet18 as the encoder architecture and investigate robustness on CIFAR10. Under the $l_\infty$ constraint, we set the maximum allowable perturbation budget $\epsilon$ to $8/255$ during both training and evaluation. Following AdvCL\cite{fan2021does}, we evaluate the learned robust representation on three metrics: Standard Accuracy (SA), Robust Accuracy (RA) and Autoattack Accuracy (AA). SA is the classification evaluated on clean examples, while RA is evaluated on adversarial examples generated by 20-step PGD attacks. AA evaluates the model accuracy under Autoattack~\cite{croce2020reliable} for mitigating the concerns for the phenomenon of obfuscated gradient. We follow~\cite{fan2021does} for the settings of SLF and AFF (see the supplementary for a detailed setup).

\subsection{Existing single-stage framework for robust SSL}

Since contrastive learning (CL) is a widely proven effective technique in SSL for representation learning without labels, for which SimCLR~\cite{chen2020simple} is a popular representative. Therefore, multiple works~\cite{jiang2020robust,kim2020adversarial,fan2021does} have adopted SimCLR as the baseline SSL method and improved its robustness by combining it with AT. Let us briefly recap how SimCLR framework works. The optimization goal of CL is to make the anchor sample be attracted close to its positive sample, \ie\ a different view augmented from the same image while being pushed away from its negative samples. The pipeline takes a batch of image samples as the input and processes it with a backbone encoder followed by a projector which is an MLP~\cite{chen2020simple}. The output is a latent vector denoted as $z$. With $\cdot$ indicating the cosine similarity between vectors and $N$ indicating the number of negative samples, the contrastive InfoNCE is shown as:
\begin{equation}
\begin{aligned}
        \mathcal{L}_{CL} &= -\log \frac{\exp({z}_a\cdot{z}_b/ \tau) }{ \exp({z}_a\cdot{z}_b/ \tau)+ \sum_{i=1}^{N}  \exp({z}_{a}\cdot{z}_{i} / \tau ) }, 
\end{aligned}
\label{eq:infonce}
\end{equation}

where $z_a$ and $z_b$ are a positive pair. $\tau$ denotes the temperature hyperparameter. The negative samples are included to prevent a collapse mode where the model outputs a constant regardless of the inputs. Note that the above loss can be simplified to a cosine similarity loss by excluding negative samples.

\textbf{RoCL.} Introducing AT to the above CL,~\cite{kim2020adversarial} is one of the pioneering works to propose a robust contrastive learning (RoCL) framework. Following the procedure in vanilla AT~\cite{madry2017towards}, RoCL first generates adversarial examples ($z_a^{adv}$) by maximizing its cosine distance from $z_b$ with multi-step PGD attacks and then updates the network by minimizing the cosine distance between all positive samples. In contrast to standard SSL, RoCL has three positive samples, $z_a$, $z_b$ and $z_a^{adv}$, which forms three contrastive losses for training the network. 

\textbf{ACL.} Concurrent to RoCL~\cite{kim2020adversarial}, another work~\cite{jiang2020robust} proposes a similar SimCLR-based approach coined as adversarial contrastive learning (ACL).~\cite{jiang2020robust} has explored to improve the robustness of SimCLR with various attempts, among which a dual stream consisting of a standard2standard (S2S) and adversarial2adversarial (A2A) performs the best. S2S is a normal CL as introduced in Eq~\ref{eq:infonce}, while A2A replaces $z_a$ and $z_b$ with adversarial examples $z_a^{adv}$ and $z_b^{adv}$ which are generated by maximizing their cosine distance to each other. 

\textbf{AdvCL.} Very recently, another SimCLR-based adversarial contrastive learning framework, which is coined as AdvCL in~\cite{fan2021does} to differentiate from ACL, has been proposed. In essence, without dual-stream design, AdvCL is more similar to RoCL than ACL. What differentiates AdvCL from them is its two distinctive components: (a) introducing another positive view which is augmented by keeping only high-frequency content; (b) adopting another supervision loss by utilizing an additional encoder pretrained on a much larger dataset (ImageNet). Empirically, these two designs improve its performance over RoCL and ACL by a large margin. As reported in~\cite{fan2021does}, this performance boost is at the cost of being three times slower than RoCL and AdvCL. If the pretraining time on ImageNet is considered, the required computation resources can be more intimidating.

\subsection{Decomposed Adversarial Contrastive Learning}
\textbf{Motivation.} Divide-and-conquer is a widely used algorithm paradigm in ML to break down a complex problem into two (or more) sub-problems which can be easier to solve. Inspired by such design philosophy, we propose to divide the complex robust SSL into two sub-problems, \ie\ (a) (non-robust) SSL and (b) (pseudo-)supervised AT. Such a decoupled optimization procedure simplifies the robust SSL by shifting the task focus from seeking an optimal strategy to combine SSL and AT to finding sub-solutions to sub-problems. Overall, with the motivation to decompose robust SSL, we propose a new two-stage framework, termed DeACL. For differentiation, we denote the encoder at stage 1 as $f_{\theta 1}$ and that at stage 2 $f_{\theta 2}$.

\textbf{Stage 1: non-robust SSL for optimizing $f_{\theta 1}$.} Following~\cite{jiang2020robust,kim2020adversarial,fan2021does}, this work mainly adopts SimCLR as the SSL method. Following~\cite{da2022solo}, we train the model for 1000 epochs. A detailed setup is listed in the supplementary. The purpose of non-robust SSL is to obtain label-alike pseudo-targets for guiding the following pseudo-supervised AT.  

\textbf{Stage 2: pseudo-supervised AT for optimizing $f_{\theta 2}$.} In vanilla supervised AT, the model training is guided by GT labels. Conceptually, the term ``label" is often associated with human predefined classes, cat or dog for instance, which do not exist in the SSL. Thus, the representation vectors obtained from SSL are termed \textit{targets} to differentiate from \textit{labels}. Moreover, since the targets are generated by a SSL pretrained model instead of human annotation, we term them \textit{pseudo-targets}. Specifically, the pseudo-targets refer to the instance-wise representation vectors by feeding the samples to a pretrained backbone encoder. They serve a similar role as GT labels to guide the supervised AT. 

\textbf{Loss design.} We use the default SSL loss (Eq~\ref{eq:infonce} for instance) to optimize $f_{\theta 1}$ at stage 1 of our DeACL. At stage 2, we optimize the encoder $f_{\theta 2}$ with the loss as:
\begin{equation}
L_{stage2}=CosSim(f_{\theta 2}(x), z_1)+ \lambda CosSim(f_{\theta 2}(x^{\text{adv}}), f_{\theta 2}(x)),
\label{eq:cossim_trades}
\end{equation}
where $CosSim$ indicates cosine similarity loss and $z_1$ indicates target vector generated from the pretrained $f_{\theta 1}$. The adversarial example $x^{\text{adv}}$ is generated by maximizing $CosSim(f_{\theta 2}(x^{\text{adv}}), z_1)$. Following AdvC~\cite{fan2021does}, 5-step PGD (with the step size $\alpha=2/255$) is adopted to generate $x^{\text{adv}}$. Eq~\ref{eq:cossim_trades} consists of two terms where the first one increases accuracy and the second one acts as a regularization loss to increase robustness. $\lambda$ is a hyper-parameter for achieving a trade-off between accuracy and robustness. In this work, we set $\lambda$ to 2 (see supplementary for the ablation study). This design is inspired by a SOTA loss in (supervised) Trades-AT~\cite{zhang2019theoretically}. A major difference from~\cite{zhang2019theoretically} is that we use $CosSim$ instead of KL divergence to measure the distance. We empirically find that $CosSim$ outperforms KL by a large margin. An alternative loss could be designed to directly minimize $CosSim(f_{\theta 2}(x^{\text{adv}}), z_1)$. Its performance is worse than Eq~\ref{eq:cossim_trades} (see the supplementary for ablation study), which aligns with the finding in prior works~\cite{zhang2019theoretically,pang2020bag}.  

\begin{figure}
         \centering
         \includegraphics[width=0.8\linewidth]{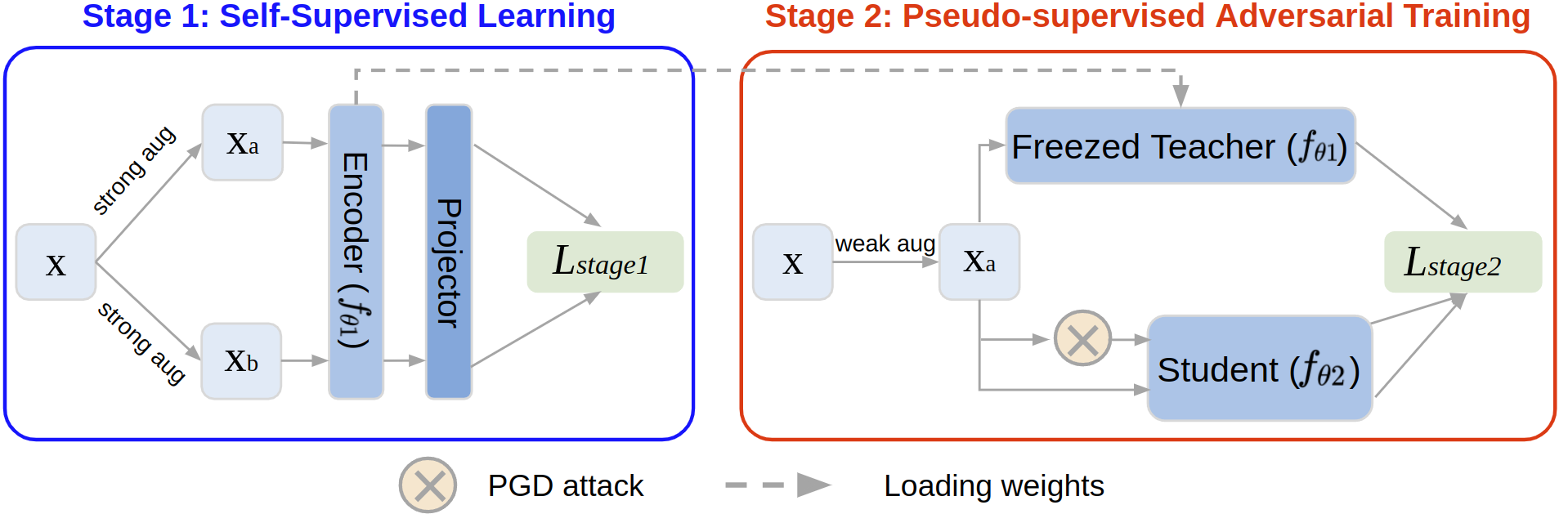}
        \caption{Overall framework of DeACL. It consists of two stages. At stage 1, DeACL performs a standard SSL to obtain a non-robust encoder. At stage 2, the pretrained encoder act as a teacher model to generate pseudo-targets for guiding a supervised AT on a student model. After two stages of training, the student model is the model of our interest.}
        \label{fig:framework}
\end{figure}

\textbf{Overall framework.} The overall framework of our DeACL is shown in Figure~\ref{fig:framework}, where the pretrained encoder in stage 1 is loaded and then frozen in stage 2 as a teacher model for generating pseudo-targets. 

% \begingroup
\begin{wraptable}[7]{ht}{0.45\textwidth}
\centering
\caption{Influence of weight initialization for the student model.}
\label{tab:weightinit}
\begin{tabular}{l|ccc}
\hline
& AA & RA & SA\\
\hline
DeACL & 45.31 &53.95 &80.17 \\
Student Scratch &44.63 &54.06 & 79.47 \\
\hline
\end{tabular}
\end{wraptable}
At stage 2, we can initialize $f_{\theta 2}$ with random weights or pretrained weights by loading $f_{\theta 1}$ to the student model. Empirically, we find that this only yields a small performance variation as shown in Table~\ref{tab:weightinit}. In the beginning, loading pretrained (non-robust) weights boosts the convergence speed (see the supplementary), which is well expected. Since this convergence and performance boost is free, our DCL by default adopts this practice.

\section{Advantages of our DeACL}
\label{sec:advantages}

\subsection{Flexible configuration for SSL and AT} \label{sec:config}

It has been shown in ~\cite{pang2020bag} that training configurations (\eg, weight decay) can have a significant influence on robustness in supervised AT. However, it is not clear how to choose the optimal configurations in robust SSL, especially considering the differences of main configurations in SSL and supervised AT, as shown in Table~\ref{tab:config_diff}. Since recent works~\cite{kim2020adversarial,jiang2020robust,fan2021does} all train the model in a single stage, it is not clear whether configurations of SSL or supervised AT should be applied. RoCL and ACL both follow the configuration as standard SSL, such as small weight decay, strong augmentation, adopting a projector head, and preventing collapse with InfoNCE loss. For the weight decay, AdvCL adopts a value that lies between those for SSL and AT. For augmentation, AdvCL adopts strong augmentation on CEs and weak augmentation on AEs.~\cite{fan2021does} shows that their AdvCL performance drops by a large margin if strong augmentation is applied on AEs. A major drawback for existing single-stage frameworks lies in seeking an optimal configuration that simultaneously fits both SSL and AT.

\begin{table}
\begin{center}
\caption{Summarization of four configuration details in various settings.}
\label{tab:config_diff}
\begin{tabular}{c|c|c|c|c|c|c|c}
\hline
\multirow{2}{*}{Configuration} &\multirow{2}{*}{SSL} &\multirow{2}{*}{AT} &\multicolumn{3}{c|}{Single-stage frameworks} &\multicolumn{2}{c}{Two-stage DeACL} \\
\cline{4-8}
& & &RoCL &ACL &AdvCL &SSL &AT \\
\hline
Weght decay & 1e-5 & 5e-4& 1e-6& 1e-6& 1e-4& 1e-5 & 5e-4 \\
Data augmentation & Strong & Weak & Strong & Strong & Strong/weak & Strong & Weak \\
Projector head & Yes  & No & Yes & Yes & Yes & Yes  & No  \\
Collapse prevention  & Yes  &  No & Yes & Yes & Yes & Yes  &  No \\
\hline
\end{tabular}
\end{center}
\end{table}

By contrast, our DeACL enables stage-specific configuration. In other words, we can freely choose the optimal experimental configurations during each stage. For the first stage which aims to train a standard model, we follow the open source SSL library~\cite{da2022solo} for optimal configuration. In the following, we detail why and how each configuration is set at stage 2 of our DeACL.

\begin{wraptable}[13]{ht}{0.42\textwidth}
\caption{Influence of weight decay at stage 2 with the SSL results on CIFAR10.}
\label{tab:weightdecay}
\begin{tabular}{c|ccc}
\hline
\multirow{2}{*}{Weight Decay} & \multicolumn{3}{c}{SLF} \\
\cline{2-4}
& AA & RA & SA\\
\hline
1e-6 &34.77 &40.45 &81.88 \\
1e-5 &36.33 &44.21 &80.29 \\
1e-4 &43.26 &52.39 &80.21 \\
5e-4 &45.31 &53.95 &80.17 \\
1e-3 &43.07 &53.32 &78.24 \\
5e-3 &24.80 &33.45 &55.76 \\
\hline
\end{tabular}
\end{wraptable}
\textbf{Weight decay.} SSL typically has a strong regularization effect due to strong augmentation and the weight decay is often set to a relatively small value, 1e-5 for instance~\cite{da2022solo}. Supervised AT, however, often suffers from robust overfitting~\cite{rice2020overfitting} and requires a large weight decay~\cite{pang2020bag}. ~\cite{pang2020bag} has performed an extensive study on a bag of tricks on supervised AT, and has found that weight decay is the most significant factor, which has a much higher influence than tricks reported in most works. With the weight decay set to 5e-4,~\cite{pang2020bag} has shown that most tricks bring no or marginal performance boost over the widely used Madry-AT~\cite{madry2017towards} and Trades-AT~\cite{zhang2019theoretically}. Table~\ref{tab:weightdecay} shows that a relatively large weight, \ie\ 5e-4, is required for achieving high robustness and accuracy. Note that ~\cite{pang2020bag} also shows that 5e-4 is the optimal weight decay for vanilla supervised AT on the same dataset CIFAR10.

\begin{wraptable}[11]{ht}{0.42\textwidth}
\caption{Influence of data augmentation at stage 2 with the SSL results on CIFAR10.}
\label{tab:augmentation}
\begin{tabular}{cc|ccc}
\hline
\multicolumn{2}{c|}{Augmentation} & \multicolumn{3}{c}{SLF} \\
\hline
AE  &CE 	&AA & RA & SA\\
\hline
Weak &Weak &45.31 &53.95 &80.17 \\
Strong &Strong &6.93 &17.43 &48.12 \\
Weak &Strong &34.47 & 45.45 &74.85 \\
Strong &Weak &7.62 &18.36 &79.93 \\
\hline
\end{tabular}
\end{wraptable}
\textbf{Data augmentation.} SSL is widely known to require strong augmentation to learn augmentation-invariant representation, while supervised training (either standard or adversarial one) typically often adopts weak augmentation. On the CIFAR10 dataset, the strong augmentation consists of random resized crop, color jittering, color change, Gaussian blur, solarization and horizontal crop, while the weak augmentation only consists of random crop (after padding) and horizontal flip. The results in \ref{tab:augmentation} show that applying strong augmentation on either clean examples (CEs) or adversarial examples (AEs) in Eq~\ref{eq:cossim_trades} yields significantly inferior performance.

\begin{wraptable}[8]{ht}{0.45\textwidth}
\centering
\caption{Influence of projector head at stage 2 with the SSL results on CIFAR10.}
\label{tab:lossprojector}
\begin{tabular}{c|ccc}
\hline
& AA & RA & SA\\
\hline
w/o projector & 45.31 &53.95 &80.17 \\
w/ Projector  &42.48 &50.60 &78.80 \\
\hline
\end{tabular}
\end{wraptable}
\textbf{Projector head.} Projector head has become a de facto standard component in SSL to be added after the backbone during training for performance boost~\cite{chen2020simple,chen2020mocov2,richemond2020byol}. After the training is done, only the backbone is kept for the downstream task. In the case of supervised AT~\cite{madry2017towards}, no projector is used during training. The results in Table~\ref{tab:lossprojector} show that adding the projector at stage 2 of our DeACL is harmful to both robustness and accuracy.

\begin{wraptable}[7]{ht}{0.58\textwidth}
\centering
\caption{Influence of collapse prevention at stage 2 with the SSL results on CIFAR10.}
\label{tab:contrastive}
\begin{tabular}{c|ccc}
\hline
& AA & RA & SA\\
\hline
w/o collapse prevention & 45.31 &53.95 &80.17 \\
w/ collapse prevention &34.18 &39.32 &72.19 \\
\hline
\end{tabular}
\end{wraptable}
\textbf{Collapse prevention.} A widely known phenomenon in SSL is that the model output a constant output, \ie\ collapse, if the loss only maximizes the cosine similarity between a pair of positive samples. A widely used approach to mitigate this phenomenon is to introduce a contrastive component, \ie\ simultaneously minimizing the cosine similarity between negative samples (See Eq~\ref{eq:infonce}). The results in Table~\ref{tab:contrastive} show that adding a contrastive component decreases the performance by a large margin. At stage 2 of our DeACL, there is low or no risk of collapse because it is supervised by distinctive pseudo-targets.

\textbf{Takeaway on the configuration.} Overall, our above investigation shows that our DeACL significantly benefits from the fact that the configurations for AT in our DeACL can be set differently from those in SSL. The best configurations at stage 2 of our DeACL are the same as those in supervised AT, which is reasonable considering the second stage our DeACL conducts a \ie\ pseudo-supervised AT. As shown in Table~\ref{table:slf}, thanks to the flexible configuration, our DeACL achieves superior performance over existing single-stage frameworks. Notably, our proposed DeACL achieves the highest robustness for both AA and RA on both CIFAR10 and CIFAR100. For SA, our DeACL outperforms all existing methods except for AdvCL on CIFAR10. On CIFAR100, our DeACL outperforms all existing methods by a large margin.

\setlength{\tabcolsep}{4pt}
\begin{table}
\begin{center}
\caption{SLF results on CIFAR10 and CIFAR100. All the methods are evaluated with ResNet18 under the same condition following~\cite{fan2021does}. We report three metrics (Auto Attack (AA), Robust Accuracy (RA), Standard Accuracy (SA)) as well as the pretraining time (in days). For all metrics, the best performance is highlighted in \textbf{bold}.}
\label{table:slf}
\begin{tabular}{c|ccc|ccc|ccc}
\hline
\multirow{2}{*}{Method}   & \multicolumn{3}{c|}{CIFAR10} & \multicolumn{3}{c|}{CIFAR100} &  \multicolumn{3}{c}{Computation resource}\\
\cline{2-10}
& AA(\%) & RA(\%) & SA(\%) & AA(\%) & RA(\%) & SA(\%) & Time&GPU&Total\\
\hline
AP-DPE   &  16.07 & 18.22  &78.30   &4.17  &6.23   &47.91 & 10.11 & 1 &10.11\\
RoCL  &23.38   &39.54   &79.90   &8.66   &18.79   &49.53  & 1.59 & 2& 3.18\\
ACL   &39.13   &42.87   &77.88   &16.33   &20.97   &47.51 & 2.65 & 1&2.65\\
AdvCL  &42.57   &50.45   &\textbf{80.85}   &19.78   &27.67   &48.34 & 3.15 & 4& 12.60\\
\hline
DeACL  & \textbf{45.31} &\textbf{53.95} &80.17 & \textbf{20.34} & \textbf{30.74} &\textbf{52.79} & {0.45} & 1& \textbf{0.45}\\
\hline
\end{tabular}
\end{center}
\end{table}
\setlength{\tabcolsep}{1.4pt}

Performing a similar investigation on existing single-stage frameworks might also bring a performance boost. However, it is not guaranteed considering the configuration trade-off between SSL and AT. Moreover, such a search for optimal configuration in single-stage frameworks can be intimidating if taking computation resources into account. In the following, we discuss another advantage of our DeACL for significantly reducing the training time.

\subsection{Two-stage DeACL is faster than single-stage frameworks}
As shown in Table~\ref{table:slf}, among the three single-stage frameworks (RoCL, ACL, AdvCL), the very recent AdvCL achieves the best robustness but at the cost of significantly more training time. As noted in~\cite{fan2021does}, their superior performance is partly attributed to introducing additional views as well as additional pseudo supervision regularization from encoder pretrained on ImageNet. These two design choices are also the reason that makes their AdvCL significantly slower. Compared with them, our DeACL requires the least training time, which is mainly attributed to the effect of disentangling SSL and AT. It is worth highlighting that our DeACL achieves a significant performance boost over RoCL and ACL, without relying on additional high-frequency views or additional supervision from ImageNet pretrained models. These two techniques might further improve the performance of our DeACL, and we leave such investigation for future work. We do not include them in this work to make our DeACL simple and fast.

\textbf{Rationale for why DeACL is fast.} Given that RoCL and ACL do not use the two design choices as AdvCL, why are they still significantly slower than our DeACL? At first sight, it seems counterintuitive that the two-stage DeACL can be faster. The rationale is briefly discussed as follows. SSL and AT are both widely known to require much longer training time than their standard supervised counterpart. Specifically, SSL often requires M (10 for instance) times more training iterations (epochs) due to the lack of GT labels. AT makes the iteration-wise training time $N$ (7 for instance) times longer because generating adversarial examples with the commonly used multi-step PGD attack is very slow. Directly solving a robust SSL requires $M\times N$ times more training time, while disentangling them into two stages is expected to only require $M+N$ times more training time. In practice, training time can be more than complex than the above reasoning rationale, depending on the implementation details.

\setlength{\tabcolsep}{4pt}
\begin{table}
\begin{center}
\caption{AFF results on CIFAR10 and CIFAR100. All the methods are evaluated with ResNet18 under the same condition following~\cite{fan2021does}. For all metrics (AA, RA, SA), our DeACL achieves the best performance which is highlighted in \textbf{bold}.}
\label{table:aff}
\begin{tabular}{c|ccc|ccc}
\hline
\multirow{2}{*}{SSL-AT}  &
\multicolumn{3}{c|}{CIFAR10} & \multicolumn{3}{c}{CIFAR100} \\
\cline{2-7}
&AA(\%) & RA(\%) & SA(\%) & AA(\%) & RA(\%) & SA(\%) \\
\hline
Supervised   
&46.19   &49.89   &79.86   &21.61   &25.86   &52.22  \\
AP-DPE     &48.13   &51.52   &81.19   &22.53  &26.89   &55.27   \\
RoCL     &47.88   &51.35   &81.01   &22.28  &27.49   &55.10  \\
ACL     &49.27   &52.82   &82.19   &23.63   &29.38   & 56.61 \\
AdvCL    &49.77   &52.77   &83.62   &24.72   &28.73   &56.77  \\
\hline
DeACL   &\textbf{50.39} &\textbf{54.18} &\textbf{83.95}   & \textbf{25.48} &\textbf{29.65} &\textbf{59.86}\\
\hline
\end{tabular}
\end{center}
\end{table}
\setlength{\tabcolsep}{1.4pt}

\section{Additional experimental results}
The results in the above section demonstrate that our DeACL outperforms existing single-stage frameworks by a large margin while requiring significantly less training time. Here, we further conduct extra experiments to verify the effectiveness of our approach from different angles.

\textbf{AFF results.} Table~\ref{table:aff} reports the AFF results on both CIFAR10 and CIFAR100. Compared to the results in Table~\ref{table:slf}, for all methods, AFF brings a consistent performance boost over SLF, which is expected since AFF also allows the encoder to be updated. Similar to the trend with SLF, we observe that our DeACL achieves SOTA performance for all the three considered metrics on both CIFAR10 and CIFAR100.

\begin{figure}
     \centering
     \begin{subfigure}{.49\textwidth}
         \centering
         \includegraphics[width=\linewidth]{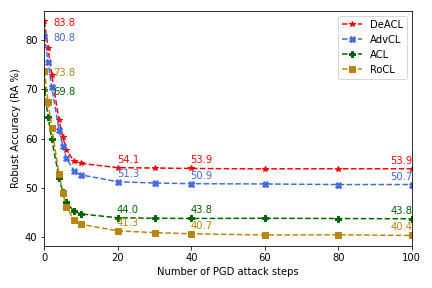}
     \end{subfigure}
     \begin{subfigure}{.49\textwidth}
         \centering
         \includegraphics[width=\linewidth]{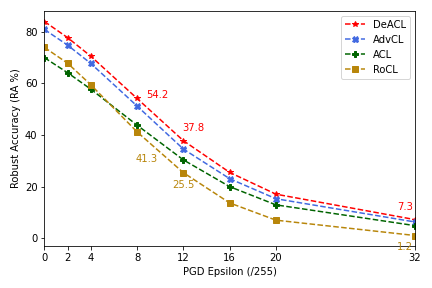}
     \end{subfigure}
        \caption{RA on CIAR10 with various PGD steps and $\epsilon$. Our DeACL consistently achieves the best performance.}
        \label{fig:viourspgdattack}
\end{figure}
\textbf{Influence of attack steps and perturbation magnitude.} For the RA, by default we use 20-step PGD, \ie\ PGD-20, with $\epsilon$ set to $l_\infty$ $8/255$. Here, we evaluate with various steps and $\epsilon$ values. The results in Figure~\ref{fig:viourspgdattack} show that our DeACL consistently outperforms existing methods by a non-trivial margin.

\begin{figure}
     \centering
     \begin{subfigure}{.24\textwidth}
         \centering
         \includegraphics[width=\linewidth]{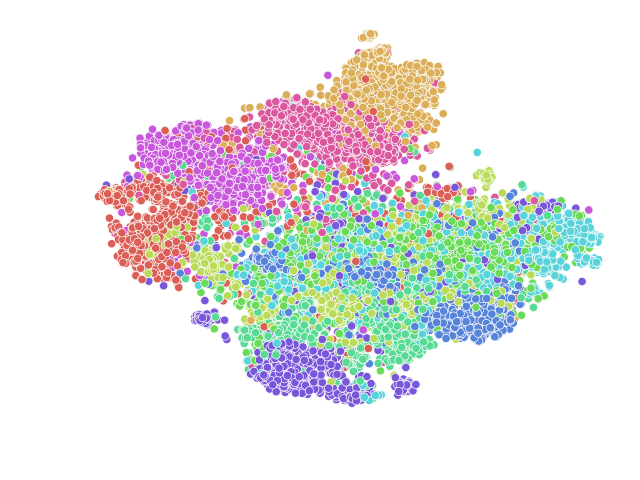}
         \caption{RoCL}
     \end{subfigure}
     \begin{subfigure}{.24\textwidth}
         \centering
         \includegraphics[width=\linewidth]{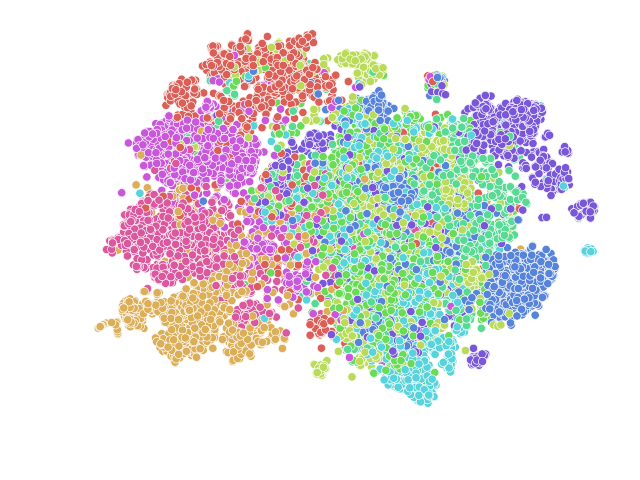}
         \caption{ACL DS}
     \end{subfigure}
     \begin{subfigure}{.24\textwidth}
         \centering
         \includegraphics[width=\linewidth]{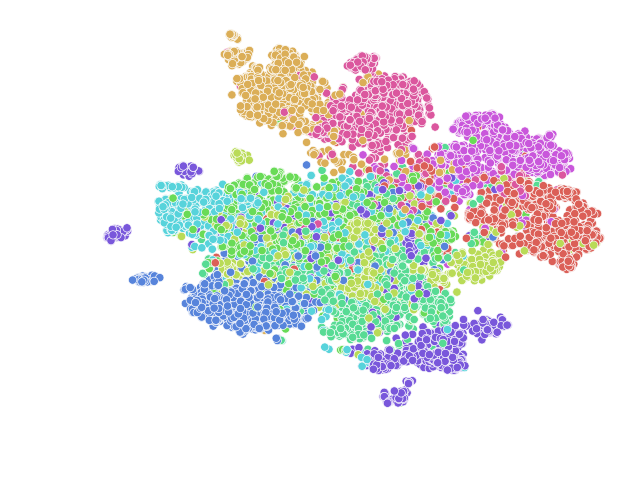}
         \caption{AdvCL}
     \end{subfigure}
     \begin{subfigure}{.24\textwidth}
         \centering
         \includegraphics[width=\linewidth]{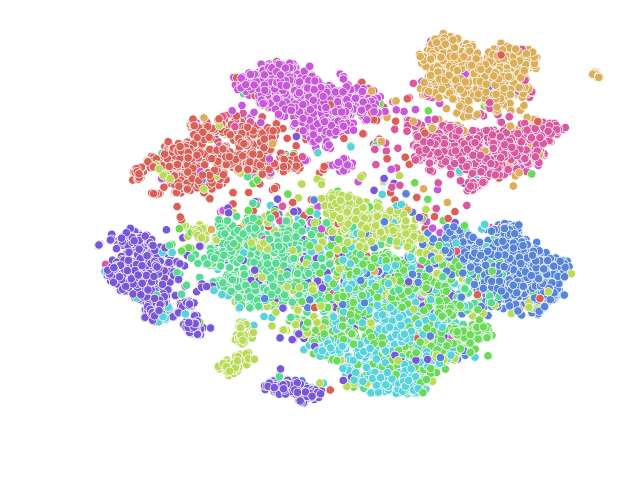}
         \caption{DeACL (ours)}
     \end{subfigure}
        \caption{Visualization of robust representation learned with different methods. Our method gives a more clear classification boundary than existing methods. }
        \label{fig:representation}
\end{figure}
\textbf{Qualitative results.} With t-SNE~\cite{van2008visualizing}, the visualization of learned representation on CIFAR10 is shown in Figure~\ref{fig:representation}, where each point is colored by its GT label. The class boundary of our DeACL is clearer than that of existing methods, which suggests that DeACL might be more robust to adversarial perturbation.

\section{Implications of our findings}

\subsection{Towards a more explainable solution}

\begin{wraptable}[12]{ht}{0.45\textwidth}
\caption{
Results on Cifar10 with various SSL frameworks at stage 1 of our DeACL.}
\label{tab:ablation1}
\begin{tabular}{c|c|c|cc}
\hline
\multirow{2}{*}{SSL frameworks} 	 & \multicolumn{3}{c}{SLF}\\\cline{2-4}
   	&AA & RA & SA\\
\hline
SimCLR~\cite{chen2020simple}  & 45.31 & 53.95 & 80.17\\
MoCo v2~\cite{chen2020mocov2}  &46.29 &53.97 & 80.56 \\
BYOL~\cite{grill2020bootstrap}  &44.14 &52.42 &80.89 \\
BarlowTwins~\cite{zbontar2021barlow}  &41.31 &50.47 &80.88  \\
VICReg~\cite{bardes2021vicreg}  &43.65 &50.56 &82.10 \\
\hline
\end{tabular}
\end{wraptable}

It is desirable to have a more explainable solution for a given task. For the task of robust SSL, the success of existing solutions based on a single-stage framework is much more difficult to explain because it couples two sub-problems. Our DeACL decouples the task, which makes the solution significantly more explainable. For example, an interesting question to ask in robust SSL is how much the SSL framework choice influences the performance. With the existing single-stage framework, such influence is much more difficult to analyze due to its interaction with AT. Note that SSL framework often has its optimal configuration setting. As we can see from Section~\ref{sec:config}, such configuration detail can have a significant influence when considering AT. With our DeACL, we can adopt a unified configuration in the second-stage AT to exclude such influence of configuration on AT. With this said, Table~\ref{tab:ablation1} reports the influence of SSL frameworks. We observe an interesting phenomenon that CL-based frameworks tend to outperform non-CL-based frameworks for achieving higher robustness but possibly at the cost of slightly lower accuracy.~\cite{xu2020adversarial} claims that adversarial momentum contrastive learning (AMOC) outperforms adversarial contrastive learning by showing a non-trivial performance gain of their AMOC over ACL. Since many configurations in MoCov2 and SimCLR are different and many other details like how to generate adversarial examples are also very different, their conclusion might be not fully convincing. Our results in Table~\ref{tab:ablation1} show that MoCov2 and SimCLR achieve comparable RA and SA, suggesting their non-trivial performance boost is likely to be caused by the influence of different configurations on AT.

\subsection{Towards a unified perspective on semi/self-supervised AT}
It is worth noting that a similar two-stage approach is also used in semi-supervised AT for exploiting unlabeled dataset. the success of such a two-stage approach in both semi-supervised and self-supervised settings suggests a new unified perspective on how to effectively exploit unlabelled dataset for learning robust representation. Intuitively, in the semi-supervised settings, the unlabeled dataset can also be used to improve the robustness in a single-stage manner through a regularization loss for instance. However,~\cite{uesato2019labels} has found that such a single-stage approach achieves inferior performance than the two-stage approach. Given that our two-stage DeACL also outperforms existing single-stage SOTA baselines in self-supervised setting, it suggests a unified understanding on semi-supervised and self-supervised AT: pseudo-targets (either pseudo-labels or pseudo-vectors) are all you need for exploiting unlabeled dataset to learn robust representation. 

Despite such a unified perspective, it is also important to note distinctions between them. Their core distinction lies in their different motivations. A semi-supervised setting allows access to labeled datasets and the motivation of using unlabeleld images is to use more samples. Note that more samples are used for training a robust model at stage 2 of semi-supervised AT than those used at stage 1 for training a non-robust model. By contrast, our DeACL in the self-supervised setting is \textit{not} motivated to increase the sample size and the second stage uses the same number of samples as those at stage 1. Instead, the motivation of the two-stage procedure in our DeACL lies in decomposing the robust SSL. Moreover, what connects the two stages is the supervision targets which are pseudo-labels and pseudo-vectors in semi-supervised and self-supervised settings, respectively. Due to this difference, at stage 2, our DeACL needs to adopt the cosine similarity loss instead of the commonly adopted CE loss. 

\section{Conclusion}
This work revisits the task of robust SSL for learning robust representation without labels. Discard the practice of seeking an optimal strategy to combine SSL and AT, we propose a novel two-stage framework termed DeACL. Our DeACL enables independent configuration for SSL and AT for achieving SOTA robustness by using significantly smaller training resources. Extensive results confirm the effectiveness and efficiency of our DeACL over existing single-stage frameworks by a significant margin. Our findings also have non-trivial implications for pushing (a) towards a more explainable solution for robust SSL and (b) towards a unified perspective of understanding on semi/self-supervised AT regarding how to effectively exploit unlabeled samples for robust representation learning. 

\noindent\textbf{Acknowledgments:}
This work was partly supported by the National Research Foundation of Korea (NRF) grant funded by the Korea government(MSIT) (No. 2022R1A2C201270611)

\clearpage
\bibliographystyle{splncs04}
\bibliography{bib_mixed}

\begin{thebibliography}{10}
\providecommand{\url}[1]{\texttt{#1}}
\providecommand{\urlprefix}{URL }
\providecommand{\doi}[1]{https://doi.org/#1}

\bibitem{athalye2018obfuscated}
Athalye, A., Carlini, N., Wagner, D.: Obfuscated gradients give a false sense
  of security: Circumventing defenses to adversarial examples. In: ICML (2018)

\bibitem{bachman2019learning}
Bachman, P., Hjelm, R.D., Buchwalter, W.: Learning representations by
  maximizing mutual information across views. NeurIPS  (2019)

\bibitem{bardes2021vicreg}
Bardes, A., Ponce, J., LeCun, Y.: Vicreg: Variance-invariance-covariance
  regularization for self-supervised learning. arXiv preprint arXiv:2105.04906
  (2021)

\bibitem{benz2020double}
Benz, P., Zhang, C., Imtiaz, T., Kweon, I.S.: Double targeted universal
  adversarial perturbations. In: ACCV (2020)

\bibitem{benz2021universal}
Benz, P., Zhang, C., Karjauv, A., Kweon, I.S.: Universal adversarial training
  with class-wise perturbations. ICME  (2021)

\bibitem{carlini2017adversarial}
Carlini, N., Wagner, D.: Adversarial examples are not easily detected. In: ACM
  Workshop on Artificial Intelligence and Security (2017)

\bibitem{carmon2019unlabeled}
Carmon, Y., Raghunathan, A., Schmidt, L., Liang, P., Duchi, J.C.: Unlabeled
  data improves adversarial robustness. NeurIPS  (2019)

\bibitem{caron2020unsupervised}
Caron, M., Misra, I., Mairal, J., Goyal, P., Bojanowski, P., Joulin, A.:
  Unsupervised learning of visual features by contrasting cluster assignments.
  arXiv preprint arXiv:2006.09882  (2020)

\bibitem{chen2020adversarial}
Chen, T., Liu, S., Chang, S., Cheng, Y., Amini, L., Wang, Z.: Adversarial
  robustness: From self-supervised pre-training to fine-tuning. In: CVPR (2020)

\bibitem{chen2020simple}
Chen, T., Kornblith, S., Norouzi, M., Hinton, G.: A simple framework for
  contrastive learning of visual representations. In: ICML (2020)

\bibitem{chen2020mocov2}
Chen, X., Fan, H., Girshick, R., He, K.: Improved baselines with momentum
  contrastive learning. arXiv preprint arXiv:2003.04297  (2020)

\bibitem{chen2021exploring}
Chen, X., He, K.: Exploring simple siamese representation learning. In: CVPR
  (2021)

\bibitem{chen2021mocov3}
Chen, X., Xie, S., He, K.: An empirical study of training self-supervised
  vision transformers. ICCV  (2021)

\bibitem{da2022solo}
da~Costa, V.G.T., Fini, E., Nabi, M., Sebe, N., Ricci, E.: Solo-learn: A
  library of self-supervised methods for visual representation learning. JMLR
  (2022)

\bibitem{croce2020reliable}
Croce, F., Hein, M.: Reliable evaluation of adversarial robustness with an
  ensemble of diverse parameter-free attacks. In: ICML (2020)

\bibitem{devlinetal2019bert}
Devlin, J., Chang, M.W., Lee, K., Toutanova, K.: {BERT}: Pre-training of deep
  bidirectional transformers for language understanding. In: Proceedings of the
  2019 Conference of the North {A}merican Chapter of the Association for
  Computational Linguistics: Human Language Technologies, Volume 1 (Long and
  Short Papers) (2019)

\bibitem{el2021xcit}
El-Nouby, A., Touvron, H., Caron, M., Bojanowski, P., Douze, M., Joulin, A.,
  Laptev, I., Neverova, N., Synnaeve, G., Verbeek, J., et~al.: Xcit:
  Cross-covariance image transformers. arXiv preprint arXiv:2106.09681  (2021)

\bibitem{ermolov2021whitening}
Ermolov, A., Siarohin, A., Sangineto, E., Sebe, N.: Whitening for
  self-supervised representation learning. In: ICML. PMLR (2021)

\bibitem{fan2021does}
Fan, L., Liu, S., Chen, P.Y., Zhang, G., Gan, C.: When does contrastive
  learning preserve adversarial robustness from pretraining to finetuning?
  NeurIPS  (2021)

\bibitem{gidaris2018unsupervised}
Gidaris, S., Singh, P., Komodakis, N.: Unsupervised representation learning by
  predicting image rotations. ICLR  (2018)

\bibitem{goodfellow2014explaining}
Goodfellow, I.J., Shlens, J., Szegedy, C.: Explaining and harnessing
  adversarial examples. In: ICLR (2015)

\bibitem{gowal2020self}
Gowal, S., Huang, P.S., van~den Oord, A., Mann, T., Kohli, P.: Self-supervised
  adversarial robustness for the low-label, high-data regime. In: ICLR (2021)

\bibitem{gowal2020uncovering}
Gowal, S., Qin, C., Uesato, J., Mann, T., Kohli, P.: Uncovering the limits of
  adversarial training against norm-bounded adversarial examples. arXiv
  preprint arXiv:2010.03593  (2020)

\bibitem{grill2020bootstrap}
Grill, J.B., Strub, F., Altch{\'e}, F., Tallec, C., Richemond, P., Buchatskaya,
  E., Doersch, C., Avila~Pires, B., Guo, Z., Gheshlaghi~Azar, M., et~al.:
  Bootstrap your own latent-a new approach to self-supervised learning.
  Advances in Neural Information Processing Systems  (2020)

\bibitem{he2019moco}
He, K., Fan, H., Wu, Y., Xie, S., Girshick, R.: Momentum contrast for
  unsupervised visual representation learning. arXiv preprint arXiv:1911.05722
  (2019)

\bibitem{he2020momentum}
He, K., Fan, H., Wu, Y., Xie, S., Girshick, R.: Momentum contrast for
  unsupervised visual representation learning. In: CVPR (2020)

\bibitem{he2016deep}
He, K., Zhang, X., Ren, S., Sun, J.: Deep residual learning for image
  recognition. In: CVPR (2016)

\bibitem{henaff2020data}
Henaff, O.: Data-efficient image recognition with contrastive predictive
  coding. In: ICML (2020)

\bibitem{hjelm2018learning}
Hjelm, R.D., Fedorov, A., Lavoie-Marchildon, S., Grewal, K., Bachman, P.,
  Trischler, A., Bengio, Y.: Learning deep representations by mutual
  information estimation and maximization. arXiv preprint arXiv:1808.06670
  (2018)

\bibitem{huang2017densely}
Huang, G., Liu, Z., Van Der~Maaten, L., Weinberger, K.Q.: Densely connected
  convolutional networks. In: CVPR (2017)

\bibitem{jiang2020robust}
Jiang, Z., Chen, T., Chen, T., Wang, Z.: Robust pre-training by adversarial
  contrastive learning. NeurIPS  (2020)

\bibitem{kim2020adversarial}
Kim, M., Tack, J., Hwang, S.J.: Adversarial self-supervised contrastive
  learning. arXiv preprint arXiv:2006.07589  (2020)

\bibitem{Lan2020ALBERT}
Lan, Z., Chen, M., Goodman, S., Gimpel, K., Sharma, P., Soricut, R.: Albert: A
  lite bert for self-supervised learning of language representations. In: ICLR
  (2020)

\bibitem{li2021esvit}
Li, C., Yang, J., Zhang, P., Gao, M., Xiao, B., Dai, X., Yuan, L., Gao, J.:
  Efficient self-supervised vision transformers for representation learning.
  arXiv preprint arXiv:2106.09785  (2021)

\bibitem{van2008visualizing}
Van~der Maaten, L., Hinton, G.: Visualizing data using t-sne. Journal of
  machine learning research  (2008)

\bibitem{madry2017towards}
Madry, A., Makelov, A., Schmidt, L., Tsipras, D., Vladu, A.: Towards deep
  learning models resistant to adversarial attacks. In: ICLR (2018)

\bibitem{moosavi2017universal}
Moosavi-Dezfooli, S.M., Fawzi, A., Fawzi, O., Frossard, P.: Universal
  adversarial perturbations. In: CVPR (2017)

\bibitem{najafi2019robustness}
Najafi, A., Maeda, S.i., Koyama, M., Miyato, T.: Robustness to adversarial
  perturbations in learning from incomplete data. NeurIPS  (2019)

\bibitem{nie2020dc}
Nie, P., Zhang, Y., Geng, X., Ramamurthy, A., Song, L., Jiang, D.: Dc-bert:
  Decoupling question and document for efficient contextual encoding. In:
  Proceedings of the 43rd International ACM SIGIR Conference on Research and
  Development in Information Retrieval (2020)

\bibitem{noroozi2016unsupervised}
Noroozi, M., Favaro, P.: Unsupervised learning of visual representations by
  solving jigsaw puzzles. In: ECCV (2016)

\bibitem{oord2018representation}
Oord, A.v.d., Li, Y., Vinyals, O.: Representation learning with contrastive
  predictive coding. arXiv preprint arXiv:1807.03748  (2018)

\bibitem{pang2020bag}
Pang, T., Yang, X., Dong, Y., Su, H., Zhu, J.: Bag of tricks for adversarial
  training. arXiv preprint arXiv:2010.00467  (2020)

\bibitem{radford2019language}
Radford, A., Wu, J., Child, R., Luan, D., Amodei, D., Sutskever, I., et~al.:
  Language models are unsupervised multitask learners. OpenAI blog  (2019)

\bibitem{rice2020overfitting}
Rice, L., Wong, E., Kolter, Z.: Overfitting in adversarially robust deep
  learning. In: ICML (2020)

\bibitem{richemond2020byol}
Richemond, P.H., Grill, J.B., Altch{\'e}, F., Tallec, C., Strub, F., Brock, A.,
  Smith, S., De, S., Pascanu, R., Piot, B., et~al.: Byol works even without
  batch statistics. arXiv preprint arXiv:2010.10241  (2020)

\bibitem{schmidt2018adversarially}
Schmidt, L., Santurkar, S., Tsipras, D., Talwar, K., Madry, A.: Adversarially
  robust generalization requires more data. In: NeurIPS (2018)

\bibitem{su2020vlbert}
Su, W., Zhu, X., Cao, Y., Li, B., Lu, L., Wei, F., Dai, J.:
  {\{}VL{\}}-{\{}bert{\}}: Pre-training of generic visual-linguistic
  representations. In: ICLR (2020)

\bibitem{szegedy2013intriguing}
Szegedy, C., Zaremba, W., Sutskever, I., Bruna, J., Erhan, D., Goodfellow, I.,
  Fergus, R.: Intriguing properties of neural networks. arXiv preprint
  arXiv:1312.6199  (2013)

\bibitem{tian2020contrastive}
Tian, Y., Krishnan, D., Isola, P.: Contrastive multiview coding. In: ECCV 2020
  (2020)

\bibitem{uesato2019labels}
Uesato, J., Alayrac, J.B., Huang, P.S., Stanforth, R., Fawzi, A., Kohli, P.:
  Are labels required for improving adversarial robustness? NeurIPS  (2019)

\bibitem{wang2020understanding}
Wang, T., Isola, P.: Understanding contrastive representation learning through
  alignment and uniformity on the hypersphere. In: ICML (2020)

\bibitem{wang2020DenseCL}
Wang, X., Zhang, R., Shen, C., Kong, T., Li, L.: Dense contrastive learning for
  self-supervised visual pre-training. In: Proc. IEEE Conf. Computer Vision and
  Pattern Recognition (CVPR) (2021)

\bibitem{wu2018unsupervised}
Wu, Z., Xiong, Y., Yu, S.X., Lin, D.: Unsupervised feature learning via
  non-parametric instance discrimination. In: CVPR (2018)

\bibitem{xie2020smooth}
Xie, C., Tan, M., Gong, B., Yuille, A., Le, Q.V.: Smooth adversarial training.
  arXiv preprint arXiv:2006.14536  (2020)

\bibitem{xie2019intriguing}
Xie, C., Yuille, A.: Intriguing properties of adversarial training at scale.
  ICLR  (2020)

\bibitem{xu2020adversarial}
Xu, C., Yang, M.: Adversarial momentum-contrastive pre-training. arXiv preprint
  arXiv:2012.13154  (2020)

\bibitem{yeh2021decoupled}
Yeh, C.H., Hong, C.Y., Hsu, Y.C., Liu, T.L., Chen, Y., LeCun, Y.: Decoupled
  contrastive learning. arXiv preprint arXiv:2110.06848  (2021)

\bibitem{zbontar2021barlow}
Zbontar, J., Jing, L., Misra, I., LeCun, Y., Deny, S.: Barlow twins:
  Self-supervised learning via redundancy reduction. ICML  (2021)

\bibitem{zhai2019adversarially}
Zhai, R., Cai, T., He, D., Dan, C., He, K., Hopcroft, J., Wang, L.:
  Adversarially robust generalization just requires more unlabeled data. arXiv
  preprint arXiv:1906.00555  (2019)

\bibitem{zhang2020resnet}
Zhang, C., Benz, P., Argaw, D.M., Lee, S., Kim, J., Rameau, F., Bazin, J.C.,
  Kweon, I.S.: Resnet or densenet? introducing dense shortcuts to resnet. In:
  WACV (2021)

\bibitem{zhang2020understanding}
Zhang, C., Benz, P., Imtiaz, T., Kweon, I.S.: Understanding adversarial
  examples from the mutual influence of images and perturbations. In: CVPR
  (2020)

\bibitem{zhang2021data}
Zhang, C., Benz, P., Karjauv, A., Kweon, I.S.: Data-free universal adversarial
  perturbation and black-box attack. In: ICCV (2021)

\bibitem{zhang2021universal}
Zhang, C., Benz, P., Karjauv, A., Kweon, I.S.: Universal adversarial
  perturbations through the lens of deep steganography: Towards a fourier
  perspective. AAAI  (2021)

\bibitem{zhang2019revisiting}
Zhang, C., Rameau, F., Lee, S., Kim, J., Benz, P., Argaw, D.M., Bazin, J.C.,
  Kweon, I.S.: Revisiting residual networks with nonlinear shortcuts. In: BMVC
  (2019)

\bibitem{zhang2022dual}
Zhang, C., Zhang, K., Pham, T.X., Yoo, C., Kweon, I.S.: Dual temperature helps
  contrastive learning without many negative samples: Towards understanding and
  simplifying moco. In: CVPR (2022)

\bibitem{zhang2022how}
Zhang, C., Zhang, K., Zhang, C., Pham, T.X., Yoo, C.D., Kweon, I.S.: How does
  simsiam avoid collapse without negative samples? a unified understanding with
  self-supervised contrastive learning. In: ICLR (2022)

\bibitem{zhang2019theoretically}
Zhang, H., Yu, Y., Jiao, J., Xing, E.P., Ghaoui, L.E., Jordan, M.I.:
  Theoretically principled trade-off between robustness and accuracy. In: ICML
  (2019)

\bibitem{zhang2019bottleneck}
Zhang, J., Han, B., Niu, G., Liu, T., Sugiyama, M.: Where is the bottleneck of
  adversarial learning with unlabeled data? arXiv preprint arXiv:1911.08696
  (2019)

\bibitem{zhuang2019local}
Zhuang, C., Zhai, A.L., Yamins, D.: Local aggregation for unsupervised learning
  of visual embeddings. In: ICCV (2019)

\end{thebibliography}
\end{document}